# EdnaML: A Declarative API and Framework for Reproducible Deep Learning

Abhijit Suprem
School of Computer Science
Georgia Institute of Technology
Atlanta, USA
asuprem@gatech.edu

Sanjyot Vaidya
School of Computer Science
Georgia Institute of Technology
Atlanta, USA

Avinash Venugopal
School of Computer Science
Georgia Institute of Technology
Atlanta, USA

Joao Eduardo Ferreira
Institute of Mathematics and Statistics
University of Sao Paulo
Sao Paulo, Brazil

Calton Pu
School of Computer Science
Georgia Institute of Technology
Atlanta, US

*Abstract*—**Machine Learning has become the bedrock of recent advances in text, image, video, and audio processing and generation. Most production systems deal with several models during deployment and training, each with a variety of tuned hyperparameters. Furthermore, data collection and processing aspects of ML pipelines are receiving increasing interest due to their importance in creating sustainable high-quality classifiers. We present EdnaML, a framework with a declarative API for reproducible deep learning. EdnaML provides low-level building blocks that can be composed manually, as well as a high-level pipeline orchestration API to automate data collection, data processing, classifier training, classifier deployment, and model monitoring. Our layered API allows users to manage ML pipelines at high-level component abstractions, while providing flexibility to modify any part of it through the building blocks. We present several examples of ML pipelines with EdnaML, including a large-scale fake news labeling and classification system with six sub-pipelines managed by EdnaML.**

*Keywords—Reproducibility, Framework, ML pipelines, provenance*

## I. INTRODUCTION

Deep Learning is the driving force behind recent advancements in a variety of fields, from image recognition, natural language processing, audio processing, video analytics, text generation, and image generation. In conjunction, a variety of tools have emerged to implement deep neural networks with optimized hardware coding for modern GPUs, such as TensorFlow [1], Caffe [2], PyTorch [3], Theano [4], CNTK [5], and Keras [6]. With increasing multidisciplinary research being conducted with deep learning, it is integral that the associated ML experiments be reproducible.

Further, as ML becomes ubiquitous in industrial processes, provenance of deployed classifiers becomes paramount for sustainable management of complex prediction systems [7]. The situation is exacerbated when considering ML pipelines, which include not only the classifiers themselves, but also data collectors, data cleaners, classifier trainers, classifier deployers, KPIs for performance monitoring with associated event triggers [8]. Compared to traditional software engineering workflows with well-defined features, interfaces, and steps, ML workflows and deployments tends towards significant experimentation in all aspects of the pipeline [7-11].

We can divide the ML pipeline into six concrete stages:

1. Data collection: Obtain data sources and push samples from sources to data stores. Sources may be authoritative or noisy; authoritative sources often provide higher quality data and, in some cases, appropriate labels. Domain knowledge can be useful to identify authoritative sources, e.g., fact checkers for fake news detection. Data sources themselves may change over time, based on new services in the world or obsolescence of existing services

2. Data preparation: Annotate a representative subset of data for training and testing. In most cases, the entire data stream from data sources may be difficult to annotate, so an annotated subset, generated either with weak supervision, label integration, or human labelers is desired.

3. Feature engineering: Identify appropriate views, transformations, or embeddings of the training data. While in some cases, a model can learn the best-fit feature engineering, hand-crafted features inject domain knowledge and can improve final accuracy

4. Model Training, Validation, and Evaluation: Models are trained and evaluated on the training/test data from Step 2. This step also includes hyperparameter tuning, AutoML for automatic training of multiple models

5. Model Deployment: The best performing model(s) are selected for deployment. If there are multiple models, weights may need to be selected based on performance, data overlap, or model confidence/intrinsic parameters

6. Model Performance Monitoring: Deployed models are monitored for performance decreases, drift, or task shift. In turn, this can lead to model updates, replacement, or retirement

The ad-hoc, experimentation-driven nature of ML pipelines means each stage is subject to a variety of tunable parameters. For example, during data collection, the URLs of data sources, collection scripts and arguments, API keys for data access, storage services and authentication keys are all potential parameters. Furthermore, parameters and metadata may be shared between stages: a data source may have specific preparation stages in, as well as specific feature engineering steps based on its content. The choice of model may also constrain the choice of transformations: transformer-based text classifiers may require subword tokenizers that exactly match the transformer classifier backbone. In addition, the parameters of the models and transformations themselves are numerous, and model sizes and complexity are only increasing.

The abundance of tunable metrics, parameters, and arguments significantly increases the complexity of a reproducible experimental setup, compared to earlier decades. To cite an example from [12], "a paper published 80 years ago on Linear Discriminant Analysis is a self-contained piece of knowledge and can be verified and reproduced with a pen and paper." On the other hand, advances in machine learning in the last decade involve datasets with millions of images and associated annotations, potential dead links in case of dehydrated datasets, hundreds of millions of model parameters, as well as several hundred hyperparameters [12]. The latter is especially striking since papers present only a few hyperparameters in their text, such as learning rate and batch size; however, the results of the experiments often depend not only on training-specific hyperparameters, but also software dependencies, configuration options, data transformations, operating system, hardware, and bare-metal workspace [13]. It is no wonder, then, that there is a reproducibility crisis in Machine Learning [7, 12, 14].

Several systems have been proposed to alleviate different aspects of provenance and reproducibility in an ML pipeline: ModelHub [15] captures variations in model architectures, ProvDB [9] manages metadata for provenance, Ground [16] provides a framework data origin management, Michelangelo [17] is a proprietary management platform in use at Uber, FBLearner [18] is a proprietary model management and storage system at Facebook, ModelDB [19] is a model management system for hyperparameter optimization, Vamsa [20] captures the principal components of a dataset that are used in a classifier, and PyTorch Lightning [21] (along with Keras [6] and fastai [22]) aims to improve experiments by reducing boilerplate code.

Many of these systems improve the productionization and reproducibility of portions of the ML pipeline: ModelHub and ModelDB focus on model parameters, and ProvDB focuses on metadata of experiments. However, very few are end-to-end systems that capture important aspects of each element in the pipeline; the relevant ones are Michelangelo, FBLearner, and SageMaker [23, 24], each of which is a proprietary system.

We believe there is a need for a provenance system for reproducible machine learning that allows both researchers and developers to manage, track, and rerun the major elements of an ML pipeline. Towards this end, such a system should rely on the following principles: (1) a bottom-up design from basic blocks to more complex pipeline operators to enable high-level and low-level design, (2) a declarative API to mimic modifications to a state-machine, and (3) extensibility and flexibility to accommodate potential new research directions.

In this paper, we present EdnaML, a declarative API and framework for deep learning to run reproducible ML experiments, with code available at ednaml.org. EdnaML is currently designed for PyTorch, as PyTorch is currently the most common language for research [25]. The rest of this paper is organized as follows: Section II covers related work in ML tracking and provenance, Section III presents the EdnaML design, Section IV covers EdnaML's implementation of pipeline abstractions in detail, and Section V describes usage and case studies.

## II. RELATED WORK

### A. Model Management

Several systems have been proposed to manage model training, deployment, and monitoring. ModelDB [19] stores already-trained models with their metadata, artifacts, and parameters for easier querying. ModelHub [15] is similar system with versioning capabilities for generated artifacts (such as intermediate stages). SageMaker [24] tracks provenance of ML experiments with a detailed schema that captures model metadata, metrics, dataset metadata, and training metadata. ProvDB [9] also presents a storage and query engine for model provenance. Vamsa [20] extends model management systems by integrating a training data query engine that connects data columns to the trained model.

### B. Data Management

Data management is as integral to ML pipelines as model management, and in many cases, more so. ML models require high quality training data to capture the task manifold, making data provenance integral to reproducibility. SQL and general database query provenance are well studied [26]; however, data provenance for machine learning is still an emerging area. Vamsa [20], mentioned earlier, records data provenance directly from python scripts. The automated metadata tracker in [8] identifies data source parameters from SparkML, scikit-learn, and MXNet experiments. ProvDB [9], MLCask [27], and other data versioning systems [28] use versioning to manage datasets, with support for external dataset versioning on git, such as DataHub [29]. DescribeML [30] presents a schema and tool for annotating dataset metadata, such as authors, tasks in data labels, version tags, and funding sources.

### C. End-to-End Management Systems

There is also significant work on proprietary ML pipeline management systems. Due to the close-source nature, it is difficult to adapt them for open research and reproducibility. Michelangelo [17] was designed by Uber to maintain ML experiments on large-scale data materialized through SparkML runtimes. FBLearner [31] uses workflow decorators with customizable input and output schemas to both track experiment parameters and generate rich UIs for easy visual reproducibility. Bighead [32] from AirBnB contains pipeline abstractions for flexible model training and rapid model development cycles. SageMaker [24] from Amazon is a full-stack management toolkit for all aspects of the ML pipeline; however, each element, from data collection, labeling, model training, evaluation, and deployment are implemented with closed-source tools. Conversely, EdnaML is an open-source end-to-end pipeline management system.

### III. EDNAML

Here, we describe the EdnaML bottom-up design, the declarative API, and provide some notes on its extensibility and flexibility.

#### A. Bottom-up design

EdnaML's bottom-up design comprises of a layered API with basic building blocks, shown in Figure 1. With the bottom-up design, we provide the basic building blocks for an ML pipeline, such as abstractions for data, model, training, and deployments. Then, we can build higher-level APIs on top of the building blocks to create a layered API that allows for provenance and tracking in the higher-level API while allowing access to the lower-level building blocks.

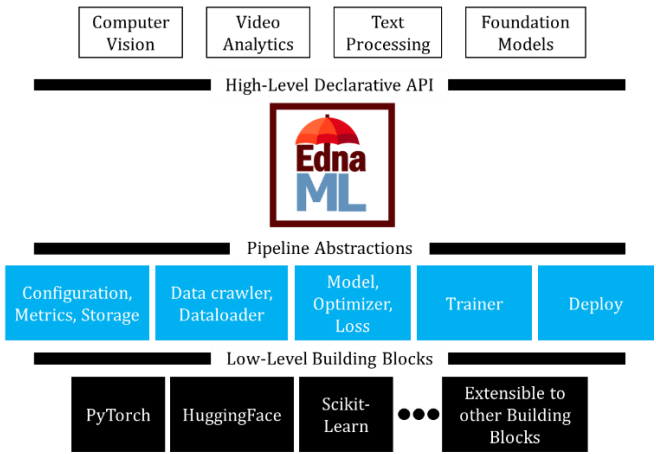

Figure 1 EdnaML's bottom-up design is a layered API that is extensible to many low-level building blocks

The lowest-level are basic building blocks comprising of existing libraries that are already heavily optimized for major machine learning tasks, such as PyTorch for model design and data loading, `HuggingFace` API for transformers and diffusers, and scikit-learn for statistical learning. We build pipeline abstractions corresponding the pipeline stages described in Sec[intro] on top of these low-level building blocks. Abstractions can be used as-is or extended to use any other building blocks the user may wish. For example, the `MODELABSTRACT` abstraction for an ML classifier (described in Section IV.C) is built on top of PyTorch's `nn.Module`; however, it can be replaced with a custom `MODELABSTRACT` on `HuggingFace`'s API as long as the `MODELABSTRACT` API is itself preserved.

Finally, we build a declarative API for instantiating, executing, reproducing, evaluating, and deploying an ML pipeline. The declarative API, accessed through the `ednaml.core.EdnaML()` class, allows for relatively easy deployment of experiments from configuration files, while still allowing access to the low-level building blocks to extend or replace any abstraction in the pipeline. We show examples of the declarative API in Section V.

#### B. Declarative API

The EdnaML declarative API simplifies the pipeline deployment process by abstracting execution steps and allowing users to focus on designing and testing pipeline components specific to their task. A pipeline is written as a configuration file specifying the desired state of each pipeline component. The pipeline can use built-in abstractions and components, or custom components. Then, EdnaML generates a pipeline to match the specifications, replacing built-in objects with custom objects, such as custom models, where applicable. When saving the state of the pipeline for reproducibility, EdnaML can then record the current state, all provided custom code, as well as any system artifacts.

Creating variations on existing experiments or deployments simply involves updating an existing configuration file with new parameters where necessary. Then, EdnaML can apply changes to the pipeline state.

By moving the complexity of pipeline design to the configuration files (the same approach as Kubernetes [33] and many modern declarative solutions such as the Twitter API [34]), we ensure a more modular design. The configuration files of an experiment can be tracked with any configuration query engine, such as NoSQL databases (e.g., MongoDB), or SQL, or git. A complete pipeline deployed through EdnaML is as simple as:

```python
from ednaml.core import EdnaML
eml = EdnaML("\path\to\config.yml")
eml.apply()
eml.train()
```

Listing 1 Code snippet showing deployment of a complete pipeline in EdnaML

#### C. Extensibility and Flexibility

While the code snippet above appears simple, we have baked significant functionality into the backend abstractions, allowing a user to completely bypass the declarative API for debugging. Further, as we will show in examples in Section V.C, each component of the pipeline can be customized; customizations are packaged with pipeline records to ensure reproducibility even with a fully bespoke pipeline.

### IV. EDNAML'S PIPELINE ABSTRACTIONS

We now describe EdnaML's configuration options and pipeline abstractions.

#### A. Configuration Options

EdnaML pipelines are described with a YAML file. A complete schema is provided in our repository at ednaml.org.

Configuration files are themselves composable, meaning multiple experiments can use the same base configuration file with experiment-specific extension file containing the differences.

```
config_base.yml                    updated_log.yml
SAVE:                              SAVE:
  MODEL_VERSION: 1                   MODEL_VERSION: 2
  MODEL_CORE_NAME: "imagenet"        LOG_BACKUP:
  MODEL_BACKBONE: "resnet18"           BACKUP: True
  MODEL_QUALIFIER: "all"               STORAGE_NAME: prometheus
  LOG_BACKUP:                          FREQUENCY: 5
    BACKUP: True
    STORAGE_NAME: log_backup
    FREQUENCY: 1
```

```python
from ednaml.core import EdnaML
eml = EdnaML(
    [ "config_base.yml",
      "updated_log.yml"
    ])
eml.apply()
eml.train()
```

Listing 2 Composable configurations allow for flexibility in designing pipelines.

We show such an example in Listing 2, where the SAVE section in a configuration base is updated with a new model version parameter, new log backup location (e.g., with a Prometheus server), and a new logging frequency. In the declarative API, this amounts to adding a list of configuration files in the appropriate order so that EdnaML can create, replace, update, or delete keys in the initial configuration.

EdnaML also contains sensible default values for most configuration options so that basic experiments can be quickly initialized for benchmarking and diagnostics, as well as to jumpstart experimentation. These built-in defaults mean basic experiments can be constructed with fewer lines, ensuring what the user focuses on in the initial stages are more meaningful operations and constructors. Furthermore, the defaults are themselves extensible and can be modified for bespoke systems that may require different default options than already provided.

```
config_base.yml                    model.py
MODEL:                             from ednaml.models import ModelAbstract
  BUILDER: ednaml                  import ednaml.core.decorators as edna
  MODEL_ARCH: CustomResnet
  MODEL_KWARGS:                    @edna.register_model
    attention: CBAM                class CustomResnet(ModelAbstract):
    layers: 80                       def __init__(self,
                                                  attention, layers...)
```

```python
from ednaml.core import EdnaML
eml = EdnaML("config_base.yml")
eml.add("model.py")
eml.apply()
eml.train()
```

Listing 3 Adding custom classes for pipeline abstractions directly through configuration options.

The configuration file serves an additional function beyond recording parameters themselves: dynamic class specification. That is, the exact model architecture, data loader, trainer, or any pipeline abstraction can be explicitly specified in the configuration file. This applies to custom implementations as well. In such cases, EdnaML first searches for the required abstraction class in any provided files, then in the built-in classes. In the configuration example in Listing 3, we have specified a model using a custom model class called CustomResnet. In the declarative API, we provide the configuration file as well as a python script containing the CustomResnet class, decorated with the appropriate register_model decorator. The decorator allows EdnaML to recognize the class is a MODELABSTRACT and import it for use with the associated configuration. During experiment recording, EdnaML will save both the configuration parameters as well as the provided file, ensuring future reproducibility with the same CustomResnet function.

### B. Datareader

The DATAREADER abstraction manages data crawling, preprocessing, and loading in batches. As such, it is itself composed of three subcomponents: Crawler, Dataset, and DATALOADER. The Dataset and DATALOADER mirror functionality to the PyTorch DATASET and DATALOADER, respectively, and are captured in a GENERATOR object. We have augmented the PyTorch data loading model with a CRAWLER that functions as the data ingest function.

EdnaML's CRAWLER abstraction takes any number of arguments as input; the arguments should contain relevant parameters as data source, any authentication tokens (either raw tokens or encrypted secrets based on organization security policy), and data source API arguments (such as dataset version), captured in the configuration's CRAWLER section. In case of a discrete experiment where a static dataset is desired, the CRAWLER generates links to data samples and passes them to the GENERATOR object. In case of a streaming process, e.g., in a deployment/serving setting or a continual learning setting, CRAWLER runs in parallel to the rest of the pipeline by periodically requesting new data from the data source. We are currently adding functionality for built-in connectors to facilitate data transfer between components with sockets.

The GENERATOR object has access, then, to either the entire dataset in case of a discrete experiment, or a socket to a stream of data in a streaming setting. The GENERATOR's DATASET object performs data preprocessing. EdnaML contains built-in Datasets for image transformations (resizing, scaling, data augmentations) through torchvision, and text transformations (tokenization, masking, padding) through HuggingFace. The DATASET object can be easily extended to custom data types and injected into a pipeline through configuration options similar to the MODELABSTRACT case in Listing 3. The DATALOADER object batches processed data for the classifier. We directly wrap around the PyTorch DATALOADER, with the same API.

### C. ModelAbstract

MODELABSTRACT is the abstract class for classifier. We extend the basic PyTorch nn.Module functionality to support more structured classifier design. To ease transition from existing architectures written in pure PyTorch, converting to MODELABSTRACT requires only rearranging code inside existing PyTorch models.

```
from ednaml.core import ModelAbstract
import ednaml.core.decorators as edna

@edna.register_model
class CustomModel(ModelAbstract):
 ❶def model_attributes_setup(self, **kwargs);
 ❷def model_setup(self, **kwargs);
 ❸def forward_impl(self, x, **kwargs);
    return logits, features, [out1, out2, ...]
```

Listing 4 MODELABSTRACT: abstraction for a classifier in EdnaML.

As we show in Listing 4, a model architecture for EdnaML that inherits from MODELABSTRACT implements 3 methods:

1. `model_attributes_setup`: Here, the class initializes any model-specific non-learnable attributes, such as embedding dimensions size or attention type. The method arguments are obtained from the `kwargs` key in the configuration's MODEL section.
2. `model_setup`: Here, the class initializes any learnable parameters, such as the layers themselves, as well as any intermediate transformations (e.g. nolinear activation layers and dropout layers). The arguments are again obtained from the `kwargs` key in the configuration's MODEL section (see Listing 3).
3. `forward_impl`: A PyTorch model's existing `forward()` function can be directly moved into `forward_impl()`, which is called by MODELABSTRACT's own `forward()` function. This allows us to perform bookkeeping and helper tasks tied to EdnaML's tracking, provenance, and plugins in `forward()`. The `forward_impl()` also needs to return a triplet consisting of logits, features, and any secondary outputs. This is also trivial to add to existing models. We are also planning functionality to automatically convert existing PyTorch model definitions into the MODELABSTRACT format.

We show in Listing 5 conversion of a conventional CNN written in PyTorch into EdnaML's MODELABSTRACT format.

### D. Model Plugins

Often there is need to update existing trained models with new functionality, or with additional execution options, such as rejection, weighting, confidence scores. Furthermore, it may be impractical to modify existing model code due to backwards compatibility, legacy requirements, or ephemerality of additional functionality. For example, reject options may need to be deployed only after input data distribution shift has occurred [35].

EdnaML also introduces a MODELPLUGIN abstraction to add additional functionality to a trained model. Plugins can be fired before or after a model sees a sample. In some cases, plugins may need training themselves, e.g. for learning model confidence or weight activations with respect to the training data for a fully-trained model. In such cases, a plugin goes through a warmup phase where it learns the required parameters by being deployed with the model on the training data again.

```
import torch.nn as nn

class CNN(nn.Module):
    def __init__(self):
        super(CNN, self).__init__()
        self.k = 2
        self.conv1 = nn.Sequential(
            nn.Conv2d(16, 32, 5, 1, self.k),
            nn.ReLU(),
            nn.MaxPool2d(self.k))
        self.out = nn.Linear(32 * 7 * 7, 10)
    def forward(self, x):
        x = self.conv1(x)
        x = x.view(x.size(0), -1)
        output = self.out(x)
        return output, x
```
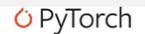

```
from ednaml.models import ModelAbstract
from torch import nn
import ednaml.core.decorators as edna

@edna.register_model()
class CNN(ModelAbstract):
    def model_attributes_setup(self, **kwargs):
        self.k = 2
    def model_setup(self, **kwargs):
        self.conv1 = nn.Sequential(
            nn.Conv2d(16, 32, 5, 1, self.k),
            nn.ReLU(),
            nn.MaxPool2d(self.k))
        self.out = nn.Linear(32 * 7 * 7, 10)
    def forward_impl(self, x, **kwargs):
        x = self.conv1(x)
        x = x.view(x.size(0), -1)
        output = self.out(x)
        return output, x, []
```

Listing 5 Converting from PyTorch to MODELABSTRACT is straightforward: model attributes and layers are simply organized under different methods.

For these cases, plugins can fire before and after an epoch of training data. These callbacks are implemented with hooks inside MODELABSTRACT's `forward()` function, and inside BASETRAINER.

We show in Listing 6 an example of a `LogitConfidence` plugin, based on a model confidence measure from [36]. This plugin needs a warmup period to learn the average confidence for each feature. During warmup, we provide the training data again so that the plugin can compute the model's average logit for each output dimension on the training data as a measure of confidence. So, we set the `activated` attribute to `False` in Line 12 during initialization and set it to `True` in the `post_epoch` hook in Line 35 after learning the necessary parameters (i.e., logit thresholds for each output dimension). Then, during deployment, we return the model outputs as well as the thresholds of logits in Lines 22-24 in the `post_forward` hook.

```
1  from ednaml.plugins import ModelPlugin
2  import ednaml.core.decorators as edna
3
4  @edna.register_plugin
5  class LogitConfidence(ModelPlugin):
6      name = "LogitConfidence"
7      def __init__(self, num_classes = 2, **kwargs):
8          super().__init__(num_classes = num_classes)
9
10     def build_plugin(self, **kwargs):
11         self.epochs = 1
12         self.activated = False
13
14     def build_params(self, **kwargs):
15         """ Set up threshold at zero, and logit tracking matrix """
16         ...
17     def post_forward(self, x, logits, feats, sec, model, **kwargs):
18         if not self.activated:
19             self.add_to_average(logits)
20             return logits, feats, sec, kwargs, {}
21         else:
22             logit, threshold = self.compute_labels(logits)
23             return logits, feats, sec,
24                 kwargs, {"logit": logit, "threshold": threshold}
25
26     def add_to_average(self, feature_logits):
27         """Perform threshold computation"""
28         ...
29
30     def post_epoch(self, model, epoch = 0, **kwargs):
31         """ Activate plugin and set threshold value"""
32         ...
33
34     def compute_labels(self, logits):
35         soft_maxes = torch.max(
36             torch.nn.functional.softmax(
37                 logits.cpu(), dim=1),dim=1)
38         return soft_maxes[0], self.logit_confidence[0,soft_maxes[1]]
39
```

Listing 6 Implementing logit-based confidence for model confidence from [36] through the EdnaML MODELPLUGIN abstraction.

### E. BaseTrainer

EdnaML formalizes the training process for ML pipelines with the BASETRAINER class. Like PyTorch Lightning, we add automate major aspects of model training, while leaving callbacks for bespoke training loops. The primary difference is we separate model training, deployment, and architecture definition into discrete components for increased composability and portability.

BASETRAINER is the basic trainer provided with EdnaML for classification. It also includes support for GAN training loops, major text-based tasks such as sentence generation, translation, and question answering by wrapping around the HuggingFace API, as well as vehicle, person, object, and face re-id. Each aspect of automation in BASETRAINER can be modified with custom trainer implementations that are packaged with the experiment for reproducibility and provenance.

BASETRAINER can perform the following tasks, reducing the boilerplate necessary for an experiment: (i) automatic gradient enabling, (ii) device-side management of GPU-to-CPU transfers and vice-versa, (ii) loss calculation and minimization (as long as built-in, default, or custom loss functions and optimizers are provided), (iii) batching plus batch-size prediction where available, (iv) backup and tracking, and (v) restarting executions if interrupted.

The BASETRAINER is enough for the majority of conventional training, e.g., image classification, prediction, translation, text classification, and GAN training. We show in Listing 7 an example of a custom trainer that implements a custom step() function to train a batch of data, and a custom evaluate_impl() function to evaluate a batch of data.

```
1  from ednaml.trainer import BaseTrainer
2  import ednaml.core.decorators as edna
3
4  @edna.register_trainer
5  class CustomTrainer(BaseTrainer):
6      def setup(self, **kwargs):
7          self.accuracy = []
8
9      def step(self, batch):
10         (samples, labels, datalabels) = batch
11         outputs = self.model(samples)
12
13         logits = outputs[0]
14         logits_loss = self.loss_fn["classification"](
15             logits=logits, labels=labels)
16         softmax_accuracy = ((logits.max(1)[1] == all_labels).float().mean())
17
18         self.losses["classification"].append(logits_loss.item())
19         self.accuracy.append(softmax_accuracy.cpu().item())
20         return logits_loss
21
22     def evaluate_impl(self):
23         logits, labels, dlabels = [],[],[]
24         with torch.no_grad():
25             for batch in self.test_loader:
26                 (samples, labels, datalabels) = batch
27                 outputs = self.model(samples)
28                 logit = outputs[0].detach().cpu()
29                 label = all_labels.detach().cpu()
30                 dlabel = all_datalabels.detach().cpu()
31
32                 logits.append(logit);labels.append(label);dlabels.append(dlabel)
33
34         logits, labels, dlabels = (cat(logits, dim=0),
35                                    cat(labels, dim=0),cat(dlabels, dim=0))
36         logit_labels = torch.argmax(logits, dim=1)
37
38         accuracy = (logit_labels == labels).sum().float() / float(labels.size(0))
39         weighted_fscore = np.mean(f1_score(labels, logit_labels, average="weighted"))
40         self.logger.info("\tAccuracy: {:.3%}".format(accuracy))
41         self.logger.info("\tWeighted F-score: {:.3f}".format(weighted_fscore))
42
43         return preds, (gt_labels, gt_datalabels), logits
```

Listing 7 A custom trainer implementation that performs additional evaluations (weighted f-score, in Line 41)) and records training accuracy from one set of labels (In Line 19)

```
1  def epoch_step(self, epoch):
2      """Trains model for an epoch."""
3      for batch in self.train_loader:
4          if self.global_batch == 0:
5              self.printOptimizerLearningRates()
6
7          self.model.train()
8          batch = self.move_to_device(batch)
9          loss: torch.Tensor = self.step(batch)
10         loss = loss / self.accumulation_steps
11         loss.backward()
12         self.accumulation_count += 1
13         if self.accumulation_count % self.accumulation_steps == 0:
14             self.updateGradients()
15             self.accumulation_count = 0
16         """
17         Intermediate-epoch saving and evaluation commented out
18         """
19         self.global_batch += 1
20
21     self.global_batch = 0
22     self.stepSchedulers()
23     self.stepLossSchedulers()
24     """
25     End-of-epoch saving and evaluation commented out
26     """
27     self.global_epoch += 1
```

Listing 8 Model training loop in BASETRAINER, in brief, with model evaluation and saving omitted.

Remaining components of BASETRAINER, such as automatic device-management, parameter optimization, and loss backpropagation are left to BASETRAINER and not re-implemented. We show BASETRAINER's training loop in brief in Listing 8; we have removed the saving and evaluation steps for brevity. The code in Listing 8's Lines 9-20 overrides the built-in step() function called in Listing 7's Line 9.

BASETRAINER performs automatic device transfers and gradient accumulation in Lines 8 and 13, respectively, and updates schedulers, if available, in Line 22.

*F. BaseDeploy*

EdnaML also provides a BASEDEPLOY class for deployments and evaluations of classifiers. Deployments do not involve training, and therefore are structurally less complex: they do not need optimizers, loss functions, schedulers for learning rates, or any training optimizations. Furthermore, we reduce technical debt by directly incorporating current states-of-the-art in model serving, such as PyTorch Serving. Planned improvements and integrations include Ray, Lightning Serve, as well as Compose for a variety of execution optimizations.

The BASEDEPLOY API is similar to BASETRAINER, in that EdnaML manages device transfers, batching, and monitoring with plugins automatically. Further, BASEDEPLOY is adaptive to non-ML workloads as well. This allows us to chain multiple pipelines for data processing with BASEDEPLOY, followed by a training pipeline with BASETRAINER, followed by a deployment with another BASEDEPLOY to create truly end-to-end projects.

We show an example of such a pipeline-of-pipelines for fake news classification in Figure 2 in Section V.C.

*G. Metrics*

Metrics are an integral part of model evaluation, tracking, and monitoring. EdnaML provides the BASEMETRIC API, with several built-in metrics; the API is, like each component, extensible to custom metrics for a project that can be packaged with an experiment for reproducibility.

```python
from ednaml.metrics import BaseMetric
import torchmetrics

class BaseTorchMetric(BaseMetric):
    def __init__(self, metric_name, metric_type,
                       torch_metric, metric_args):
        super().__init__(metric_name, metric_type)
        self.metric = torch_metric
        self.metric_args = metric_args
        self.metric_obj = None
        self.results = None

    def build_module(self, **kwargs):
        self.metric_obj = self.metric(**self.metric_args) \
            if self.metric_args else self.metric()
        self.results = []

    def update(self, **kwargs):
        result = self.metric_obj(**kwargs)
        self.results.append(result)

import torchmetrics.KLDivergence as Torch_KLDivergence
KLDivergence = BaseTorchMetric(
    metric_name='Pytorch_KL_Divergence',
    metric_type='Relative Entropy',
    torch_metric = Torch_KLDivergence
)
```

Listing 9 `torchmetric` wrapper in EdnaML, plus KLDivergence metric implementation using the wrapper

We also provide a wrapper around `torchmetrics`, which is a popular library for ML metrics implemented for PyTorch. We provide the basic wrapper implementation for most metrics in `torchmetrics` in Listing 9; the last few lines show how to instantiate the KLDivergence metric from `torchmetrics` directly in EdnaML.

*H. Storage*

EdnaML's STORAGE API allows users to define storage backends for data ingest, saving, and pipeline backups for reproducibility. The default option is file-based storage, where any pipeline outputs are saved to a directory in the execution path. EdnaML currently supports local storage, remote network attached storage, Azure blobs, and GCP Cloud Storage. We have also used custom plugins in existing experiments for SQL, MongoDB, and other NoSQL backends in an ad-hoc basis. In some cases, we need non-traditional storage options: for example, `mlflow` is a well-known machine learning tracking API that records experiment metrics. In conjunction with BASEMETRIC, we wrote STORAGE plugins to record pipeline metadata and metrics in an `mlflow` backend store. We plan to add these, as well as S3 support to the release branch.

EdnaML also formalizes the pipeline tracking backend by setting distinct options for storing the components of a pipeline. Inside the configuration, users can specify distinct backends, storage frequency and other storage-specific parameters to save: configuration, logs, models, model artifacts, model plugins, and metrics. Pipeline outputs can be saved with custom Storage solutions or built-in objects through a BASETRAINER or BASEDEPLOY object.

V. USAGE AND CASE STUDIES

We now cover some examples of EdnaML's usage, beginning with the traditional MNIST example.

*A. MNIST Example*

The configuration for a basic MNIST experiment using Resnet-18 is provided in Listing 10. Compared to more recent ML code managers and frameworks, we do not abstract all control: instead, we simply move it to configuration files that are easier to read and manage. We have not shown default options in this configuration file; a complete example is on our repository. We can control the entire experiment from here; thus, we can have multiple versions of this file that tweak different parts of the pipeline, like the normalization for MNIST in Line 20, or the learning rate in Line 36. Given this configuration file, the pipeline code is shown in in Listing 11.

*B. MNIST with Modified Parameters*

Now we show a second experiment where we modified some parameters of the original experiment. While the original configuration file can be used, we could also apply changes to the desired state with a second configuration overriding the first. This allows for more modular experiment design, with a base configuration file, plus additional pipeline component-specific changes. The snippet in Listing 12 exists in a second configuration file that can be executed in conjunction with the configuration in Listing 10 with the code in Listing 11. The modified configuration is now at MODEL_VERSION of 2, with a `BASE_LR` of 1e-4 (compared to 1e-3, see Line 36 in Listing 10), and training for 10 epochs.

```
 1  EXECUTION:
 2    EPOCHS: 5
 3    TRAINER: ClassificationTrainer
 4    TRAINER_ARGS:
 5      accumulation_steps: 2
 6  DATAREADER:
 7    DATAREADER: TorchvisionDatareader
 8    GENERATOR_ARGS:
 9      tv_dataset: MNIST
10      tv_args: {root: "Data/", args: {download: true}}
11    DATASET_ARGS:
12      label_name: mnist_digits
13  SAVE:
14    MODEL_CORE_NAME: mnist_resnet
15    BACKUP: False
16    SAVE_FREQUENCY: 5
17  TRANSFORMATION:
18    ARGS:
19      i_shape: [28,28]
20      normalization: [0.1307, 0.3081, 0.5]
21      channels: 1
22  TRAIN_TRANSFORMATION:
23    ARGS:
24      h_flip: 0.5
25  MODEL:
26    BUILDER: ednaml_model_builder
27    MODEL_ARCH: ClassificationResnet
28    MODEL_BASE: resnet18
29    MODEL_KWARGS: {initial_channels: 1}
30  LOSS:
31    - LOSSES: ['SoftmaxLogitsLoss']
32      LAMBDAS: [1.0]
33      LABEL: mnist_digits
34  OPTIMIZER:
35    - OPTIMIZER: Adam
36      BASE_LR: 1.0e-3
```

Listing 10 Configuration for a basic MNIST experiment using Resnet-18

```
from ednaml.core import EdnaML
eml = EdnaML("\path\to\config.yml")
eml.apply()
eml.train()
```

Listing 11 Code for executing the MNIST pipeline

```
EXECUTION:                          from ednaml.core import EdnaML
  EPOCHS: 10                        eml = EdnaML(["config1.yml",
SAVE:                                              "config2.yml"])
  MODEL_VERSION: 2                  eml.apply()
OPTIMIZER:                          eml.train()
  - OPTIMIZER: Adam
    BASE_LR: 1.0e-4
```

Listing 12 Modifying Listing 10 with updated parameters and injecting them into the pipeline.

### C. Pipeline-in-Pipelines with EdnaML

Most ML experiments or deployments require multiple pipelines to manage data transformations. Furthermore, intermediate transformations may be used in other models, pipelines, or services. EdnaML allows for a unified approach to such problems. Here, we describe using EdnaML to manage various aspects of a COVID-19 fake news labeling and training pipeline for Twitter data, shown in Figure 2.

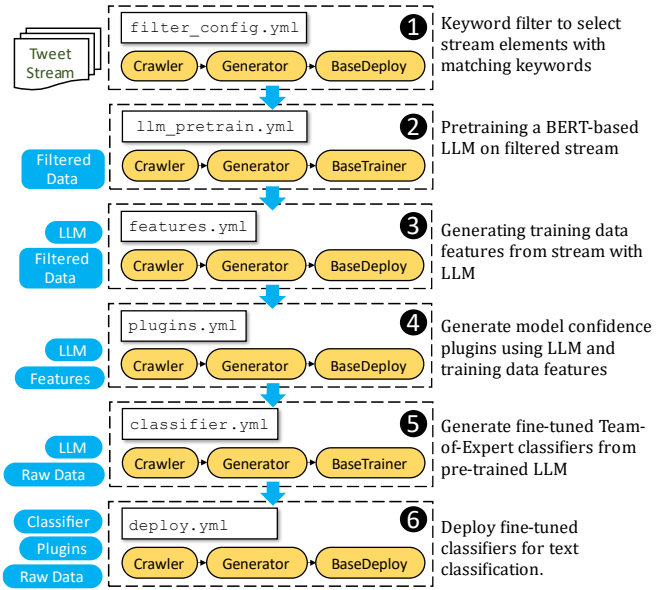

Figure 2 We use EdnaML to manage a large-scale labeling and training pipeline for Twitter-based fake news data.

Here, each sub-pipeline is launched from its respective configuration file (while we have shown a single configuration file in each pipeline; in reality, these are composable configurations in the same vein as in Listing 12, allowing for dynamic adjustment within the entire system):

1. The first pipeline performs keyword filtering on the tweets from the Twitter Sampled Stream API with a `BASEDEPLOY` object, and saves periodically to a network attached storage (NAS)
2. Next, we periodically launch the second pipeline (frequency managed by EdnaML) to pre-train a large-language model (LLM) [37] on the filtered data.
3. The pre-trained LLM is provided new training data in the form of samples from authoritative sources [38] such as news articles about misinformation, and generates features that are saved to a NAS
4. Next, we build plugins for the LLM to perform training data clustering compute embedding smoothness [39].
5. We then generate fine-tuned classifiers from the pre-trained LLM from Step 2 using authoritative labeled data plus weak labels from prior fine-tuned classifiers.
6. Finally, we can deploy fine-tuned classifiers with EdnaML for live predictions, when desired.

## VI. CONCLUSIONS

Experimentation is integral to ML pipeline; furthermore, many aspects of ML experimentation are non-deterministic, from initial parameters of models to training loops to dropout layers that stochastically update weight paths, to random batch selection. As such, repeatable and reproducible experiments and models are integral. In this paper, we have presented EdnaML, our framework with a declarative API to run and maintain reproducible ML experiments. We have provided an alpha release at ednaml.org, with several examples. We covered EdnaML's bottom-up design principle, where we began with basic building blocks and added layers of abstraction that improve provenance, tracking, maintenance, and reproducibility of machine learning pipelines. We also provided details on EdnaML's major components, with several code examples, as well as a case study of a large-scale labeling and classification system for fake news detection. We hope EdnaML aids the research community in furthering open-source end-to-end pipeline management systems.

ACKNOWLEDGMENT

This research has been partially funded by National Science Foundation by CISE/CNS (1550379, 2026945, 2039653), SaTC (1564097), SBE/HNDS (2024320) programs, and gifts, grants, or contracts from Fujitsu, HP, and Georgia Tech Foundation through the John P. Imlay, Jr. Chair endowment. Any opinions, findings, and conclusions or recommendations expressed in this material are those of the author(s) and do not necessarily reflect the views of the National Science Foundation or other funding agencies and companies mentioned above.